\documentclass[final,5p,times,twocolumn]{elsarticle}

\usepackage{amssymb}
\usepackage{amsmath}
\usepackage{amssymb}
\usepackage{amsmath}
\usepackage{times}  
\usepackage{helvet}  
\usepackage{courier}  
\usepackage[hyphens]{url}  
\usepackage{graphicx} 
\usepackage{natbib}  
\usepackage{caption} 
\usepackage{url}  
\usepackage{hyperref}
\usepackage{amsthm}
\usepackage{pifont}
\usepackage{booktabs}
\usepackage{longtable}
\usepackage{makecell}
\usepackage{listings}
\usepackage{multirow}
\usepackage{pdflscape}
\usepackage{epstopdf}
\usepackage{mathtools}
\usepackage{indentfirst}
\usepackage{subcaption} 
\usepackage{algorithm}
\usepackage{algorithmic}

\usepackage{newfloat}
\usepackage{listings}
\DeclareCaptionStyle{ruled}{labelfont=normalfont,labelsep=colon,strut=off} 
\floatstyle{ruled}
\newfloat{listing}{tb}{lst}{}
\floatname{listing}{Listing}

\newcommand{\model}{PViT\xspace}
\usepackage[dvipsnames,table]{xcolor}

\usepackage[labelsep=period]{caption}
\renewcommand{\paragraph}[1]{\vspace{0.2em}\noindent \textbf{#1\hspace{0.2em}}}

\usepackage{xspace} 

\usepackage{pifont} 

\usepackage{amsmath, amssymb} 
\usepackage{dsfont} 

\def\d1{\mathds{1}} 

\DeclareMathOperator*{\argmax}{arg\,max}

\def\1{\mathbf{1}}

\journal{arxiv}

\begin{document}

\begin{frontmatter}

\title{\model: Prior-augmented Vision Transformer for Out-of-distribution Detection}

\author[1]{Tianhao Zhang}
\author[2]{Zhixiang Chen}
\author[1]{Lyudmila S. Mihaylova}

\affiliation[1]{organization={School of Electrical and Electronic Enigeering, The University of Sheffield}, 
            city={Sheffield},
            country={United Kingdom}}

\affiliation[2]{organization={School of Computer Science, The University of Sheffield}, 
            city={Sheffield},
            country={United Kingdom}}
\begin{abstract}
Vision Transformers (ViTs) have achieved remarkable success over various vision tasks, yet their robustness against data distribution shifts and inherent inductive biases remain underexplored. To enhance the robustness of ViT models for image Out-of-Distribution (OOD) detection, we introduce a novel and generic framework named Prior-augmented Vision Transformer (\model). Taking as input the prior class logits from a pretrained model, we train PViT to predict the class logits. During inference, \model identifies OOD samples by quantifying the divergence between the predicted class logits and the prior logits obtained from pre-trained models. Unlike existing state-of-the-art(SOTA) OOD detection methods, \model shapes the decision boundary between ID and OOD by utilizing the proposed prior guided confidence, without requiring additional data modeling, generation methods, or structural modifications. Extensive experiments on the large-scale \textsc{ImageNet} benchmark, evaluated against over seven OOD datasets, demonstrate that \model significantly outperforms existing SOTA OOD detection methods in terms of FPR95 and AUROC. The codebase is publicly available at \url{https://github.com/RanchoGoose/PViT}.
\end{abstract}




\end{frontmatter}

\section{Introduction}\label{sec:intro}
In recent years, the Transformer~\cite{vaswani2017attention}, characterized by its innovative attention mechanism, has achieved a significant success in various domains, extending its success from natural language processing to various vision tasks. The inception of the Vision Transformer (ViT)~\cite{dosovitskiy2020image} represents a pivotal moment in the adaptation of Transformer architectures for vision applications, setting the stage for subsequent models that exhibit remarkable performance enhancements through increased depth and scale, albeit at the cost of heightened computational demands~\cite{kirillov2023segany}. However, the exploration into enhancing Out-of-Distribution (OOD) detection within these architectures has lagged, especially when compared to the extensive research conducted for Convolutional Neural Networks (CNNs)-based models~\cite{ren2019likelihood, serra2019input, xiao2020likelihood, pimentel2014review, denouden2018improving, zong2018deep}.  

OOD detection is a crucial machine learning technique that aims to identify test samples from distributions divergent from the training data distribution. This technique is essential for differentiating between inputs that are part of the training distribution and those that are not~\cite{yang2021generalized}. The importance of proficient OOD detection is underscored in safety-critical real-world deployments, where encountering novel classes is inevitable. Addressing the generalization and OOD detection capabilities of ViTs becomes imperative \cite{paul2021vision}.
%
\begin{figure*}[t]
    \centering
    \includegraphics[width=0.7\linewidth]{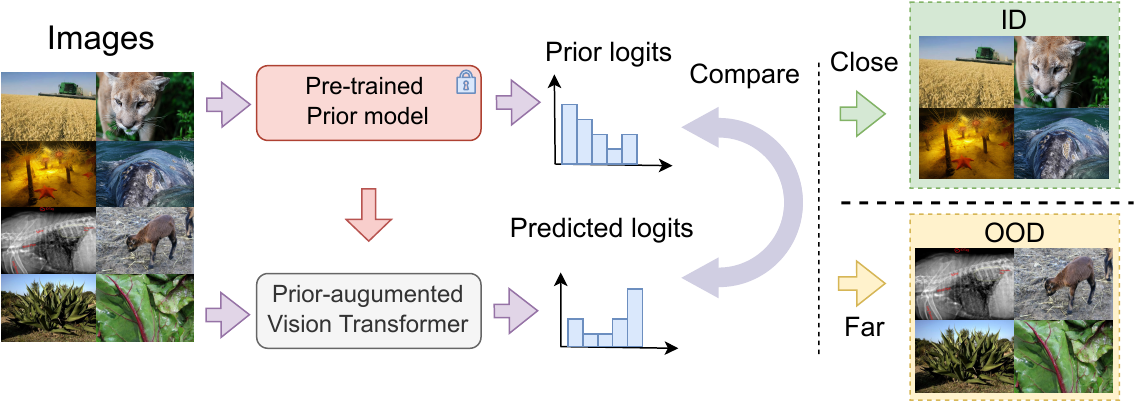}
    \caption{A brief overview of the proposed method. The ID images are taken from \textsc{IMAGENET-1K}, and the OOD images are sourced from \textsc{Openimage-O}. The distinction between ID and OOD images is made by measuring the difference between the prior logits and the predicted logits.}
    \label{fig:breif}
\end{figure*}

We explore the strategic incorporation of prior knowledge in vision models to enhance their safety-related capabilities. The hypothesis is that, like humans, AI models can benefit from contextual cues, improving their ability to accurately identify and classify data. This suggests that prior knowledge may be crucial in helping models discern nuanced data distributions, leading to our central research question:

\begin{center}\textbf{Will the model enhance the performance in OOD detection if provided with prior knowledge?}\end{center}

In this work, our goal is to leverage the advantages of ViTs to design a scalable solution that improves their robustness and performance for OOD detection. Compared to traditional CNNs, ViTs directly operate on sequences of image patches~\cite{han2020survey}, producing results through the \textit{attention} mechanism. ViTs are effective at capturing long-range dependencies between patches but often neglect local feature extraction, as 2D patches are projected into vectors using a simple linear layer. Some recent studies have begun to focus on enhancing the modeling capacity for local information~\cite{liu2021swin}. Enabling ViTs to incorporate additional contextual data broadens their analytical scope beyond the immediate visual input, creating a strategic opportunity to integrate prior knowledge derived from high-performing pre-trained vision models into the learning process.

Following the idea of integrating the prior knowledeg into a ViT, we propose the novel Prior-augmented Vision Transformer (\model) for OOD detection. As illustrated in Fig.~\ref{fig:breif}, \model is designed to generate predictions that closely align with the prior logits for In-Distribution (ID) data while exhibiting significant divergence for OOD data. The prior knowledge is derived from a pre-trained model on the ID dataset, referred to as the \textit{prior model} in this paper. \model is trained using the prior predictions generated by the prior model. During inference, \model employs the proposed \textit{Prior Guide Energy} (PGE) score to effectively distinguish OOD instances by quantifying the divergence between the prior logits and the predicted logits.




We demonstrate that our proposed framework, \model, is highly effective for OOD detection, particularly on large-scale datasets such as \textsc{Imagenet}. Compared to state-of-the-art OOD detection methods, \model achieves remarkable performance improvements, reducing the FPR95 by up to 20\% and increasing the AUROC by up to 7\% compared to the best baseline. Additionally, \model eliminates the need for generating synthetic outlier data while maintaining high accuracy on ID datasets. 

The key contributions of this paper are summarized as follows:

\begin{enumerate}
    \item We introduce \model, a novel and generic framework that integrates prior knowledge into ViT, thereby enhancing model robustness and OOD detection capabilities.
    \item We introduce the Prior Guide Energy as an effective scoring method for OOD detection by measuring the similarity between the prior class logits and the predicted class. 
    \item We conduct comprehensive experiments on various benchmarks across a diverse set of ID and OOD datasets, providing qualitative analyses of \model and offering insightful discussions on the impact of incorporating prior knowledge into ViT models.
\end{enumerate}

The remainder of this paper is structured as follows. Section~\ref{sec:background} reviews related work and provides the necessary background for our study. Section~\ref{sec:method} presents a detailed description of our proposed \model framework, including its architecture and the novel prior-guided OOD scoring mechanism. In Section~\ref{sec:experiments}, we evaluate the performance of \model through extensive experiments on large-scale OOD benchmarks, including \textsc{CIFAR} and \textsc{IMAGENET-1K}. Section~\ref{sec:scale} presents comprehensive ablation studies with both quantitative and qualitative analyses. Section~\ref{sec:diss} discusses the implications of our approach, highlights its current limitations, and outlines potential directions for future research. Finally, Section~\ref{sec:conclusion} summarizes the main results and contributions. The \emph{Appendices} provide additional details on the datasets and the models used in this paper to ensure reproducibility. 


\section{Related Works} \label{sec:background}
\subsection{Out-of-Distribution (OOD) Detection}
Deep learning models are often overconfident when classifying samples from different semantic distributions, leading to inappropriate predictions in tasks such as image classification and text categorization~\cite{hendrycks2016baseline}. This issue has prompted the emergence of the field of OOD detection, which requires models to reject inputs that are semantically different from the training distribution and should not be predicted by the model. OOD detection is a critical area of research aimed at ensuring the safe deployment of AI systems~\cite{LIU2025110947}.  Over the years, various methods for OOD detection have emerged, broadly categorized into techniques focused on network modifications and score-based approaches to distinguish between ID and OOD samples in the embeddings or latent feature spaces~\cite{odin18iclr}.

Methods modifying network behavior often employ techniques like truncation. For example, ODIN~\cite{odin18iclr} perturbs the input using gradient vectors to amplify detection scores, while ReAct~\cite{sun2021react} applies thresholding to clip hidden layer activations. These approaches enhance the network’s ability to separate ID and OOD samples.

Score-based methods involve developing scalar metrics to quantify the likelihood of a sample being OOD. Classifier-based approaches, often referred to as \textit{confidence scoring}, leverage the neural network's classification layer. A seminal work in this domain is the Maximal Softmax Probability (MSP) method, which serves as a baseline for OOD detection~\cite{hendrycks2016baseline}. Subsequent advancements include the energy function~\cite{liu2020energy}, which provides a bias-free estimation of class-conditional probabilities, and the maximum-of-logit technique~\cite{vaze2021open, hendrycks2019scaling, dhamija2018reducing}, which combines class likelihood with feature magnitude for improved performance. Additionally, Kullback–Leibler (KL) divergence has been used to compare predictions with a uniform distribution, enhancing class-dependent information~\cite{hendrycks2019scaling}.

Distance-based methods form another key category of OOD detection, identifying samples based on their spatial relationship to ID data in the feature space. The Mahalanobis detector~\cite{lee2018simple} computes distances to class-wise means with shared feature covariance, while SSD assumes a single Gaussian distribution for ID samples~\cite{sehwag2021ssd}. Non-parametric methods like k-Nearest Neighbors (k-NN) offer precise boundary delineation and have been improved through NNGuide, which enhances differentiation in distant datasets~\cite{sun2022out, park2023nearest}. 


Other popular OOD detection methods include enhancing the model robustness by creating \emph{outliers}, also referred to as outlier exposure approaches. These methods impose a strong assumption on the availability of OOD training data, which can be infeasible in practice. When no OOD samples are available, some methods attempt to synthesize OOD samples to enable ID/OOD separability. Existing works leverage GANs to generate OOD training samples and force the model predictions to be uniform~\cite{lee2018training}, generate boundary samples in the low-density region~\cite{DBLP:journals/corr/abs-1910-04241}, or produce high-confidence OOD samples~\cite{confgan18nipsw}. However, synthesizing images in the high-dimensional pixel space can be difficult to optimize. Recent work, VOS~\cite{du2022towards}, proposed synthesizing virtual outliers from the low-likelihood region in the feature space, which is more tractable given the lower dimensionality. In object detection, \cite{du2022unknown} proposes synthesizing unknown objects from videos in the wild using spatial-temporal unknown distillation. Recent advances focus on localizing OOD regions in complex visual environments, such as urban driving scenarios~\cite{du2022siren, ming2022delving, du2023dream}. Such outlier exposure methods often require additional training and generating the synthesized data, which reduces scalability and adaptability. Compared to existing state-of-the-art methods, our proposed \model distinguishes itself by eschewing reliance on synthesized data or external outliers for training, thereby enhancing its scalability and adaptability across diverse frameworks~\cite{DBLP:journals/corr/abs-2103-02603, du2023dream}.

\subsection{Vision Transformers}
Originally proposed for machine translation, Transformers have ascended to the state-of-the-art in numerous Natural Language Processing (NLP) tasks~\cite{vaswani2017attention}. The vanilla ViT~\cite{dosovitskiy2020image}, representing the first adaptation of a purely Transformer-based model for image classification, has shown competitive performance with state-of-the-art CNNs. Like their NLP counterparts, ViTs lack the local receptive fields and weight-sharing properties of CNNs. Instead, they use positional encodings and self-attention to capture positional relationships. This flexibility allows ViTs to learn any data relationship, but they may require more data to learn patterns that CNNs naturally capture due to their spatial hierarchy and locality biases~\cite{10088164}.

Following the paradigm of ViT showing superior results on remarkable performance, a series of variants of ViT have been proposed to improve the performance on a variety of visual tasks, such as image classification~\cite{touvron2021training, singh2022revisiting, ZHANG2024109979, DING2023109532, liu2021swin}, image segmentation~\cite{zhang2024morphoseg, GUO2024110491}, and object detection~\cite{carion2020end, HUA2025111111, XIE2024110172}. DeiT~\cite{touvron2021training}, also known as Data-efficient image transformer, is later proposed as a competitive convolution-free transformer by training on only the ImageNet database. Swin Transformers~\cite{liu2021swin,dong2021cswin} performs local attention within a window and introduces a shifted window partitioning approach for cross-window connections.


Other than pure vision tasks, Transformers has also been used for Bayesian inference. A recent study exploring the use of Transformers for Bayesian Inference~\cite{müller2023transformers} has broadened the scope of their applicability. This research demonstrates that when trained on prior samples, Transformers are capable of effectively approximating the posterior predictive distribution (PPD), even in scenarios involving small tabular datasets~\cite{hollmann2023tabpfn}. In contrast, our method is designed to work with large-scale image data, showcasing the versatility of taking advatages of prior information in handling diverse data scales and types.

While the introduction of specialized tokens in ViTs is a relatively unexplored area~\cite{huang2022vision, park2022matteformer}, our work pioneers the use of a \textit{prior token} for OOD detection. The concept of a \textit{prior token} is not first introduced by us, as seen in applications like MatteFormer~\cite{park2022matteformer} for image matting, which integrates trimap information via a Prior-Attentive Swin Transformer block. Our approach, however, diverges significantly as it repurposes this concept for enhancing OOD detection in ViTs.

\section{Methodology} \label{sec:method}
\begin{figure*}[t]
    \centering
    \includegraphics[width= 0.8\textwidth]{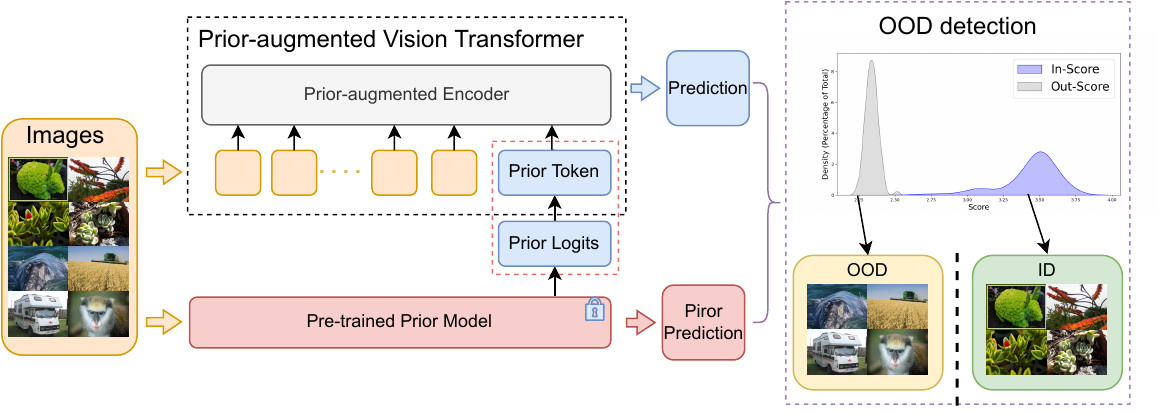} 
    \caption{Framework of our proposed \model. During the training stage, \model processes the ID image patches \( \mathcal{D}_{\text{in}}^{\text{train}} \) alongside the prior token \( \mathbf{T}_{\text{prior}} \), which embeds prior knowledge from the pre-trained prior model. During testing, the prior model \( \theta_{\text{prior}} \) continues to provide the prior logits for the OOD data \( \mathcal{D}_{\text{out}}^{\text{test}} \) to \model. The predicted class logits are then used to calculate the prior-guided OOD score, enabling the differentiation between ID and OOD data. Other components, including position embeddings, the classification (cls) token, and the flattening of image patches, follow the implementation of the vanilla ViT~\cite{dosovitskiy2020image}.}
    \label{fig:method}
\end{figure*}

This section lays the foundation of our approach, starting with the problem setup (Section.~\ref{sec:preli}) to establish the necessary background. We then offer a comprehensive overview of our \model presented in Section.~\ref{sec:overview}, detailing its architecture and functionality with Figure.~\ref{fig:method}. Finally, we explore our OOD scoring mechanism, particularly emphasizing the role of the PGE score in differentiating OOD instances in Section.~\ref{sec:PCE}.

\subsection{Preliminaries} \label{sec:preli}
In the context of image classification, let \( \mathcal{X} = \mathbb{R}^d \) represent the input space, and \( \mathcal{Y} = \{1, \ldots, K\} \) denote the finite set of labels for \( K \) classes. The training dataset \( \mathcal{D}_{\text{in}}^{\text{train}} = \{(\mathbf{x}_i, y_i)\}_{i=1}^n \) consists of pairs \( (\mathbf{x}_i, y_i) \), where a classification function \( f_\theta: \mathcal{X} \rightarrow \mathbb{R}^K \) predicts class scores. The predicted label \( \widehat{y} \) is obtained as \( \widehat{y} = \argmax_{k} f_\theta^{(k)}(\mathbf{x}) \), corresponding to the class with the highest score.

For testing on unseen data, the objective is to train a model capable of distinguishing OOD inputs \( \mathcal{D}_{\text{out}}^{\text{test}} = \{(\mathbf{x}_j, y_j)\}_{j=1}^m \), where labels \( y_j \) do not belong to \( \mathcal{Y} \). To achieve this, a binary classification approach, also referred to as the decision rule for OOD detection, is employed:

\begin{equation}
    \mathbf{x} = \left\{
    \begin{array}{ll}
        \text{ID} & \text{if } S(\mathbf{x};\theta) \geq \gamma, \\
        \text{OOD} & \text{if } S(\mathbf{x};\theta) < \gamma,
    \end{array}
    \right.
    \label{eq:ood_rule}
\end{equation}

\noindent where the threshold \( \gamma \) is selected to ensure high classification accuracy for ID data, typically set at 95\%. The score \( S(\mathbf{x};\theta) \), also known as 'confidence', represents the classifier-based detection score.

\subsection{Prior-augmented Vision Transformer (PViT)} \label{sec:overview}
The architecture of the Prior-augmented Vision Transformer (PViT) is depicted in Fig.~\ref{fig:method}. The implementation of PViT follows the foundational structure of the vanilla Vision Transformer (ViT)~\cite{dosovitskiy2020image}. In the conventional ViT, an input image $\mathbf{x} \in \mathbb{R}^{H \times W \times C}$ is transformed into a sequence of flattened 2D patches $\mathbf{x}_p \in \mathbb{R}^{N \times (P^2 \cdot C)}$. Here, $(H, W)$ represents the resolution of the original image, $C$ denotes the number of channels, $(P, P)$ defines the resolution of each patch, and $N = \frac{HW}{P^2}$ indicates the total number of patches, which effectively determines the sequence length for the Transformer.

Similar to the vanilla ViT, a learnable embedding $\mathbf{z}_0^0$, initially set to $\mathbf{x}_{\text{class}}$, serves as the class embedding. This embedding is designed to capture the global image representation and is iteratively updated throughout the Transformer layers. After $L$ layers of the Transformer encoder, its final state, $\mathbf{z}_L^0 \in \mathbb{R}^{1 \times D}$, serves as the aggregated image representation, denoted by $\mathbf{y} \in \mathbb{R}^{D}$, where $D$ is the dimensionality of the embedding space. Given an input image, the model first computes patch embeddings $\mathbf{E}_{\text{patches}} \in \mathbb{R}^{N \times D}$, where $N$ represents the number of patches. A class token $\mathbf{t}_{\text{cls}} \in \mathbb{R}^{1 \times D}$ is then prepended to this sequence of embeddings. Positional encodings $\mathbf{P}_{\text{pos}} \in \mathbb{R}^{(N+1) \times D}$ are added to provide the sequence with spatial information, resulting in the final sequence of embeddings $\mathbf{E}_{\text{pos}} = [\mathbf{t}_{\text{cls}}; \mathbf{E}_{\text{patches}}] + \mathbf{P}_{\text{pos}}$.

\vspace{2mm}
\noindent\textbf{Prior Token Integration.} Given a pre-trained \textit{prior model} parameterized by \( \theta_{\text{prior}} \), the prior logits vector \( \mathbf{p} \in \mathbb{R}^{K} \) represents the classification output logits, serving as prior knowledge for \model. To incorporate this prior knowledge into the ViT architecture, we introduce a special token, termed the \textit{prior token}. This token, \( \mathbf{t}_{\text{prior}} \in \mathbb{R}^{1 \times D} \), encapsulates the prior knowledge and is input alongside the patch tokens and the class token into the prior-augmented encoder, where it is processed by the attention mechanism. 

To create the prior token, the logits vector \( \mathbf{p} \in \mathbb{R}^{K} \) from the pre-trained classifier is first normalized using the softmax function. These normalized logits are then projected into the embedding dimension \( D \) to form the prior token \( \mathbf{t}_{\text{prior}} \):
\begin{equation}
\mathbf{t}_{\text{prior}} = \mathbf{W}_{\text{proj}} \cdot \text{softmax}(\mathbf{p}),
\label{eq:token}
\end{equation}

\noindent where $\mathbf{W}_{\text{proj}} \in \mathbb{R}^{D \times C}$ is a learnable projection matrix designed to transform the class-wise priors into the embedding space, aligning them dimensionally with the patch embeddings.

The prior token is then scaled by a factor \( \alpha \in \mathbb{R} \), a hyperparameter that modulates the influence of prior knowledge. This scaling balances the model's attention between the prior token and the image-derived embeddings, optimizing overall performance. The scaled prior token \( \mathbf{t}_{\text{prior}} \) is then replicated across the batch, resulting in:

\begin{equation}
\mathbf{T}_{\text{prior}} = \alpha \cdot \mathbf{t}_{\text{prior}} \otimes \mathbf{1}_{B \times 1 \times D},
\label{eq:rep}
\end{equation}

\noindent where \( B \) denotes the batch size, and \( \otimes \) represents the outer product with a vector of ones, effectively broadcasting \( \mathbf{t}_{\text{prior}} \) across the batch. The batch-level prior token \( \mathbf{T}_{\text{prior}} \) is appended to the positionally encoded sequence, forming the complete input \( \mathbf{T} = [\mathbf{E}_{\text{pos}}; \mathbf{T}_{\text{prior}}] \) for the encoder.

The concatenated sequence $\mathbf{T}$ is processed through the Transformer encoder layers to yield the final representations. Our model follows the architecture of the vanilla ViT~\cite{dosovitskiy2020image}, employing multi-headed self-attention (MSA)~\cite{vaswani2017attention} (Eq.~\eqref{eq:MSA}) and multi-layer perceptron (MLP) blocks (Eq.~\eqref{eq:MLP}). Layer normalization (LN) is applied before each block (Eq.~\eqref{eq:final_representation}), as described in the following equations:
\begin{align}
    \mathbf{Z}_0 &= \mathbf{T}, \label{eq:start}\\
    \mathbf{Z}^\prime_{\ell} &= \text{MSA}(\text{LN}(\mathbf{Z}_{\ell-1})) + \mathbf{Z}_{\ell-1}, \quad \ell = 1 \ldots L, \label{eq:MSA} \\ 
    \mathbf{Z}_{\ell} &= \text{MLP}(\text{LN}(\mathbf{Z}^\prime_{\ell})) + \mathbf{Z}^\prime_{\ell}, \quad \ell = 1 \ldots L,  \label{eq:MLP} \\
    \mathbf{Y} &= \text{LN}(\mathbf{Z}_L[0]),  \label{eq:final_representation}
\end{align}

\noindent where $\mathbf{Z}_L[0]$ represents the final layer's class token representation. The output $\mathbf{Y} \in \mathbb{R}^{D}$ serves as the input to a classifier head for the task at hand.

In the context of image classification, the primary training objective for \model is to minimize the divergence between the model's predicted distribution and the true label distribution. The overall training objective is achieved through the minimization of the cross-entropy loss function~\( \mathcal{L}_\text{CE} \), which is formulated as:

\begin{equation}
    \mathcal{L}_\text{CE} = -\sum_{i=1}^K \log P_{\text{\model}}(y_i | x_i, \mathcal{D}, \pi; \theta),
    \label{eq:loss}
\end{equation}

\noindent where \( \theta \) represents the parameters of \model, \( y_i \) the true labels, \( x_i \) the input data, \( \mathcal{D} \) the dataset, and \( \pi \) the prior information. 

\subsection{Prior Guide Energy for OOD Detection} \label{sec:PCE}
Given a base confidence score function \( S_{\text{base}}(\mathbf{x}; \theta) \), we propose a Prior Guide Energy (PGE) method to effectively differentiate between ID and OOD data by incorporating prior knowledge:

\begin{equation}
    S_{\text{PGE}}(\mathbf{x}; \theta) = S_{\text{base}}(\mathbf{x}; \theta) \cdot G(\mathbf{x}; \theta, \theta_{\text{prior}}),
    \label{eq:overall_score}
\end{equation}

\noindent where \( G(\mathbf{x}; \theta, \theta_{\text{prior}}) \) is the \textit{guidance term}, designed to measure the similarity between the prior embeddings and the outputs of \model. 

\noindent\textbf{Energy as a Base Confidence Score}
The base confidence score \( S_{\text{base}}(\mathbf{x}; \theta) \) is derived from the \textit{Energy} score~\cite{liu2020energy}, defined as:
\begin{equation}
    E(\mathbf{x}; \theta) = -\log \sum_{i=1}^{K} e^{f_i(\mathbf{x}; \theta)}, 
    \label{eq:energy}
\end{equation}

\noindent where \( f_i(\mathbf{x}; \theta) \) denotes the logit corresponding to class \( i \) output by the model. The effectiveness of the energy score for OOD detection arises from its relationship with the model's learned representation. The energy score is particularly effective for OOD detection due to the following properties:
\begin{itemize}
    \item \textbf{Push-Pull Dynamics}: During training with the negative log-likelihood (NLL) loss, the energy of the correct label is minimized while the energies of incorrect labels are increased, creating a sharp confidence boundary between ID and OOD examples.
    \item \textbf{Free Energy Interpretation}: The energy score implicitly incorporates the \textit{Free Energ} of the system (log partition function), allowing it to model the overall uncertainty of predictions across all classes.
    \item \textbf{Non-probabilistic Efficiency}: The score is efficiently computed via the \texttt{logsumexp} operator, making it computationally advantageous compared to probabilistic density estimates.
\end{itemize}

For ID data, the logits \( f_i(\mathbf{x}; \theta) \) form a concentrated distribution, resulting in low energy scores. Conversely, for OOD data, the logits tend to distribute more uniformly across all classes, leading to higher energy scores. This separation is a natural consequence of the NLL training objective, which explicitly pushes down the energy for ID data while increasing the energy for irrelevant classes. To align with the conventional OOD detection definition, we using the negative energy score, \( -E(\mathbf{x}; \theta) \), as \( S_{\text{base}}(\mathbf{x}; \theta) \).

\noindent\textbf{Prior Guidance Term}
While the energy score is a powerful standalone confidence measure, it can benefit from additional prior information to enhance its discriminative ability. To this end, we introduce the guidance term \( G(\mathbf{x}; \theta, \theta_{\text{prior}}) \), which evaluates the similarity between prior knowledge and the predictions of the current model. Here we introduce one of the possible options: Cross Entropy (CE). CE is a widely used metric for quantifying the cost of matching multi-class predictions, and it is recognized as an effective and direct method for defining training targets in classification tasks. As demonstrated by the results presented in Section~\ref{sec:experiments}, CE emerges as the optimal guidance term by utilizing the prior logits and the predicted class as inputs:

\begin{equation}
    G(\mathbf{x}; \theta, \theta_{\text{prior}}) = -\sum_{i=1}^{K} y_i \log\left(q_i(\mathbf{x}; \theta_{\text{prior}})\right),
    \label{eq:guidance_term}
\end{equation}

\noindent where \( q_i(\mathbf{x}; \theta_{\text{prior}}) \) is the probability of class \( i \) based on the prior logits, and \( y_i \) is the predicted class by \model. This guidance term measures the dissimilarity between the model’s predictions and the prior distribution. A higher cross entropy score indicates greater alignment with the prior distribution, suggesting that the data are likely ID, while a lower score suggests potential OOD data. Other optional guidance terms for measuring similarity than CE are discussed in the Section~\ref{sec:scale}.

\noindent\textbf{Overall PGE Score} 
By combining the base confidence score and the prior guidance term, the PGE score is defined as:

\begin{align}
    S_{\text{PGE}}(\mathbf{x}; \theta) = \underbrace{S_{\text{energy}}(\mathbf{x}; \theta)}_{\uparrow~\text{for in-distribution $\mathbf{x}$}} \cdot \underbrace{G(\mathbf{x}; \theta, \theta_{\text{prior}})}_{\uparrow~\text{for in-distribution $\mathbf{x}$}}.
    \label{eq:spge}
\end{align}

The guidance term amplifies the base confidence score, resulting in a higher overall PGE score for ID data. Conversely, OOD data are characterized by lower PGE scores. By setting an appropriate threshold \( \gamma \), the PGE score can effectively separate ID and OOD data:

\begin{align}
\begin{cases}
S_{\text{PGE}}(\mathbf{x}; \theta) > \gamma, & \text{if } \mathbf{x} \in \mathcal{D}_{\text{in}}, \\
S_{\text{PGE}}(\mathbf{x}; \theta) \leq \gamma, & \text{if } \mathbf{x} \in \mathcal{D}_{\text{out}}.
\end{cases}
\end{align}

The next section presents a detailed description of the datasets, evaluation metrics and the overall evaluation of the proposed OOD detection approach.

\section{Experiments} \label{sec:experiments}
\noindent\textbf{Datasets.} To assess the model performance, we use the small-scale \textsc{Cifar}~\cite{cifar} and the large-scale \textsc{Imagenet-1K}~\cite{deng2009imagenet} dataset, as our ID training datasets. \textsc{Cifar-10} and \textsc{Cifar-100} are used interchangeably as ID and OOD datasets due to their similarities yet distinct characteristics. We use the standard train/validation/test splits for training and testing. In main results reported in Tab.~\ref{tab:vit_imagenet} and Tab.~\ref{tab:swag_imagenet} where \textsc{Imagenet-1K} is used as ID data, we employ a range of natural image datasets as OOD benchmarks, including \textsc{iNaturalist}~\cite{van2018inaturalist}, \textsc{Sun} \cite{xiao2010sun}, \textsc{Textures}~\cite{cimpoi2014describing}, \textsc{Places} \cite{zhou2017places}, \textsc{NINCO}~\cite{bitterwolf2023ninco}, \textsc{OpenImage-O}~\cite{haoqi2022vim}, and \textsc{SSB-hard}~\cite{vaze2022openset}. In Tab..~\ref{tab:cifar100} where \textsc{Cifar-100} is used as ID dataset, the following OOD test datasets are used to text the OOD performance of \model: \textsc{Cifar-10}~\cite{cifar}, \textsc{Textures}~\cite{cimpoi2014describing},  
\textsc{Places} \cite{zhou2017places}, \textsc{Lsun}~\cite{DBLP:journals/corr/YuZSSX15}, \textsc{iSun}~\cite{DBLP:journals/corr/XuEZFKX15}, and  \textsc{Svhn}~\cite{netzer2011reading}. 

\noindent\textbf{Training Details.} \label{sec:training_details}
The \model is trained with a configuration that includes a hidden dimension of \(384\), a depth of \(12\) layers, \(6\) MSA heads, and a MLP dimension of \(768\). The Adam optimizer~\cite{kingma2014adam} is used with hyperparameters \(\beta_1 = 0.9\) and \(\beta_2 = 0.999\) over \(20\) epochs. Training begins with an initial learning rate of \(0.1\) and employs a batch size of \(256\), a momentum of \(0.9\), and a weight decay of \(1 \times 10^{-3}\). A linear learning rate decay schedule is applied after \(5\) warm-up epochs. For different ID datasets, we utilize various prior models, including ResNet~\cite{he2016deep} and ViT models~\cite{dosovitskiy2020image} along with their variants~\cite{touvron2021training, singh2022revisiting}. For ImageNet-1K as the ID dataset, the ViT models are pre-trained on \textsc{ImageNet-21K} and subsequently fine-tuned on \textsc{ImageNet-1K}. All pre-trained models used as prior models are publicly available. 
\begin{table*}[ht]
    \centering
    \scalebox{0.55}{

\begin{tabular}{lcccccccccccccccc}
\toprule
OOD Datasets                                               & \multicolumn{2}{c}{\textsc{inaturalist}}             & \multicolumn{2}{c}{\textsc{sun}}                     & \multicolumn{2}{c}{\textsc{places}}                  & \multicolumn{2}{c}{\textsc{textures}}                & \multicolumn{2}{c}{\textsc{ninco}}                   & \multicolumn{2}{c}{\textsc{openimage\_o}}            & \multicolumn{2}{c}{\textsc{ssb-hard}}                & \multicolumn{2}{c}{Mean}                                          \\
\cmidrule(lr){2-3} \cmidrule(lr){4-5} \cmidrule(lr){6-7} \cmidrule(lr){8-9} \cmidrule(lr){10-11} \cmidrule(lr){12-13} \cmidrule(lr){14-15} \cmidrule(lr){16-17}
Metrics                                                    & \small{FPR95}                            & \small{AUROC}                          & \small{FPR95}                            & \small{AUROC}                          & \small{FPR95}                            & \small{AUROC}                          & \small{FPR95}                            & \small{AUROC}                          & \small{FPR95}                            & \small{AUROC}                          & \small{FPR95}                            & \small{AUROC}                          & \small{FPR95}                            & \small{AUROC}                          & \small{FPR95}                            & \small{AUROC}                          \\
Methods                                                    & \multicolumn{1}{l}{\small{$\downarrow$}} & \multicolumn{1}{l}{\small{$\uparrow$}} & \multicolumn{1}{l}{\small{$\downarrow$}} & \multicolumn{1}{l}{\small{$\uparrow$}} & \multicolumn{1}{l}{\small{$\downarrow$}} & \multicolumn{1}{l}{\small{$\uparrow$}} & \multicolumn{1}{l}{\small{$\downarrow$}} & \multicolumn{1}{l}{\small{$\uparrow$}} & \multicolumn{1}{l}{\small{$\downarrow$}} & \multicolumn{1}{l}{\small{$\uparrow$}} & \multicolumn{1}{l}{\small{$\downarrow$}} & \multicolumn{1}{l}{\small{$\uparrow$}} & \multicolumn{1}{l}{\small{$\downarrow$}} & \multicolumn{1}{l}{\small{$\uparrow$}} & \multicolumn{1}{l}{\small{$\downarrow$}} & \multicolumn{1}{l}{\small{$\uparrow$}} \\ 
\midrule
MSP~\cite{hendrycks2016baseline}     & 15.33                            & 97.19                          & 46.48                            & 90.12                          & 53.34                            & 87.89                          & 42.82                            & 89.61                          & 51.58                            & 88.31                          & 27.85                            & 94.98                          & 67.41                            & 80.98                          & 43.55                            & 89.87                          \\
MaxLogit~\cite{hendrycks2019scaling} & 36.96                            & 94.00                          & 77.03                            & 81.56                          & 80.82                            & 78.65                          & 65.20                            & 84.01                          & 81.15                            & 75.57                          & 68.05                            & 86.29                          & 92.28                            & 63.62                          & 71.64                            & 80.53                          \\
Energy~\cite{liu2020energy}          & 31.18                            & 92.64                          & 61.58                            & 82.66                          & 64.99                            & 81.83                          & 59.61                            & 81.07                          & 63.80                            & 82.82                          & 44.86                            & 89.03                          & 79.21                            & 73.03                          & 57.89                            & 83.30                          \\
SSD~\cite{sehwag2021ssd}             & 18.17                            & 96.65                          & 50.50                            & 89.36                          & 55.97                            & 87.35                          & 47.07                            & 88.69                          & 54.12                            & 87.88                          & 31.41                            & 94.31                          & 71.02                            & 80.27                          & 46.89                            & 89.22                          \\
ViM~\cite{wang2022vim}               & 70.91                            & 80.81                          & 93.42                            & 60.38                          & 95.42                            & 56.20                          & 85.46                            & 69.19                          & 92.85                            & 53.42                          & 87.02                            & 64.36                          & 92.98                            & 50.81                          & 88.30                            & 62.17                          \\
KNN~\cite{sun2022out}                & 78.14                            & 74.68                          & 91.20                            & 59.32                          & 94.04                            & 55.43                          & 89.34                            & 63.16                          & 93.33                            & 52.17                          & 91.21                            & 59.42                          & 94.61                            & 48.49                          & 90.27                            & 58.95                          \\
NNGuide~\cite{park2023nearest}       & 81.44                            & 76.72                          & 94.97                            & 51.18                          & 95.88                            & 49.40                          & 85.20                            & 68.99                          & 94.85                            & 55.80                          & 92.75                            & 60.46                          & 96.71                            & 43.87                          & 91.68                            & 58.06                          \\
\rowcolor[HTML]{EFEFEF} 
PViT + CE                                                 & \textbf{13.08}                   & \textbf{97.56}                 & 41.56                            & 90.99                          & 49.97                            & 88.40                          & \textbf{39.56}                   & \textbf{90.43}                 & \textbf{49.20}                   & \textbf{88.33}                 & \textbf{25.62}                   & \textbf{95.26}                 & 61.31                            & \textbf{82.19}                 & \textbf{40.04}                   & \textbf{90.45}                 \\
\rowcolor[HTML]{EFEFEF} 
PViT + ED                                                  & 18.03                            & 96.18                          & \textbf{28.75}                   & \textbf{91.85}                 & \textbf{37.01}                   & \textbf{90.24}                 & 78.01                            & 79.42                          & 51.02                            & 87.08                          & 49.12                            & 88.92                          & \textbf{60.23}                   & 80.86                          & 46.02                            & 87.79                         \\ 
\bottomrule
\end{tabular}%
}
    \caption{OOD detection results for \model using the original ViT-B/16 as the prior model. All benchmarks are evaluated on the same prior model with an ID accuracy of \textbf{75.73\%}, while PViT achieves an ID accuracy of \textbf{76.61\%}. OOD detection results are reported for \textsc{IMAGENET-1K} as the ID dataset. Arrows (\(\uparrow\) and \(\downarrow\)) indicate that larger or smaller values are better, respectively. All values are percentages. \textbf{Bold} numbers indicate superior results. \textbf{CE} refers to using cross entropy as the prior guidance, and \textbf{ED} refers to using Euclidean Distance as the prior guidance.}
    \label{tab:vit_imagenet}
\end{table*}

\begin{table*}
    \centering
    \scalebox{0.55}{

\begin{tabular}{lcccccccccccccccc}
\toprule
{OOD Datasets}                      & \multicolumn{2}{c}{\textsc{inaturalist}} & \multicolumn{2}{c}{\textsc{sun}} & \multicolumn{2}{c}{\textsc{places}} & \multicolumn{2}{c}{\textsc{textures}} & \multicolumn{2}{c}{\textsc{ninco}} & \multicolumn{2}{c}{\textsc{openimage\_o}} & \multicolumn{2}{c}{\textsc{SSB-hard}} & \multicolumn{2}{c}{Mean} \\
\cmidrule(lr){2-3} \cmidrule(lr){4-5} \cmidrule(lr){6-7} \cmidrule(lr){8-9} \cmidrule(lr){10-11} \cmidrule(lr){12-13} \cmidrule(lr){14-15} \cmidrule(lr){16-17}
Metrics                                       & \small{FPR95}                            & \small{AUROC}                          & \small{FPR95}                            & \small{AUROC}                          & \small{FPR95}                            & \small{AUROC}                          & \small{FPR95}                            & \small{AUROC}                          & \small{FPR95}                            & \small{AUROC}                          & \small{FPR95}                            & \small{AUROC}                          & \small{FPR95}                            & \small{AUROC}                          & \small{FPR95}                            & \small{AUROC}                          \\
Methods                                       & \multicolumn{1}{l}{\small{$\downarrow$}} & \multicolumn{1}{l}{\small{$\uparrow$}} & \multicolumn{1}{l}{\small{$\downarrow$}} & \multicolumn{1}{l}{\small{$\uparrow$}} & \multicolumn{1}{l}{\small{$\downarrow$}} & \multicolumn{1}{l}{\small{$\uparrow$}} & \multicolumn{1}{l}{\small{$\downarrow$}} & \multicolumn{1}{l}{\small{$\uparrow$}} & \multicolumn{1}{l}{\small{$\downarrow$}} & \multicolumn{1}{l}{\small{$\uparrow$}} & \multicolumn{1}{l}{\small{$\downarrow$}} & \multicolumn{1}{l}{\small{$\uparrow$}} & \multicolumn{1}{l}{\small{$\downarrow$}} & \multicolumn{1}{l}{\small{$\uparrow$}} & \multicolumn{1}{l}{\small{$\downarrow$}} & \multicolumn{1}{l}{\small{$\uparrow$}} \\ 
\midrule
MSP      & 52.21                                  & 88.05                                 & 67.13                                  & 80.75                                & 68.99                                  & 80.17                                & 62.39                                  & 82.03                                & 61.10                                  & 84.37                                & 85.50                                  & 68.48                                & 74.04                                  & 77.98                                & 67.34                                  & 80.26                                \\
MaxLogit & 52.90                                  & 85.33                                 & 67.57                                  & 76.36                                & 69.66                                  & 75.16                                & 58.79                                  & 80.76                                & 60.34                                  & 81.17                                & 86.23                                  & 64.18                                & 74.04                                  & 72.33                                & 67.08                                  & 76.47                                \\
Mahalanobis     & 21.40                                  & 95.55                                 & 62.72                                  & 84.65                                & 63.08                                  & 83.73                                & 53.07                                  & 87.00                                & 43.62                                  & \textbf{91.66}                       & 76.52                                  & 70.48                                & 58.59                                  & \textbf{86.23}                       & 54.14                                  & 85.61                                \\
Energy          & 62.68                                  & 79.98                                 & 73.14                                  & 70.53                                & 74.55                                  & 68.81                                & 60.09                                  & 78.29                                & 65.66                                  & 76.45                                & 88.06                                  & 59.24                                & 78.85                                  & 65.76                                & 71.86                                  & 71.29                                \\
SSD             & 25.93                                  & 94.14                                 & 68.80                                  & 80.31                                & 68.98                                  & 79.49                                & 50.32                                  & 85.20                                & 47.37                                  & 90.17                                & 80.19                                  & 66.61                                & 61.47                                  & 83.46                                & 57.58                                  & 82.77                                \\
ViM             & 18.67                                  & 96.20                                 & 61.65                                  & 82.44                                & 63.21                                  & 80.61                                & 45.43                                  & 88.10                                & 42.53                                  & 91.88                                & 76.66                                  & 68.88                                & 55.58                                  & 85.52                                & 51.96                                  & 84.80                                \\
KNN             & 68.61                                  & 85.77                                 & 80.21                                  & 79.21                                & 78.06                                  & 78.31                                & 67.66                                  & 82.31                                & 65.77                                  & 85.59                                & 89.65                                  & 59.09                                & 77.65                                  & 76.93                                & 75.37                                  & 78.17                                \\
NNGuide         & 65.39                                  & 88.63                                 & 78.40                                  & 81.60                                & 76.15                                  & 80.56                                & 64.41                                  & 85.02                                & 64.72                                  & 87.66                                & 88.92                                  & 63.27                                & 76.45                                  & 80.35                                & 73.49                                  & 81.01                                \\
\rowcolor[HTML]{EFEFEF} PViT + CE    & \textbf{2.25}                          & \textbf{99.39}                        & 30.99                                  & 93.38                                & 40.46                                  & 90.97                                & \textbf{32.94}                         & \textbf{92.08}                       & 41.51                                  & 90.77                                & 31.29                                  & 91.59                                & 58.27                                  & 84.53                                & 33.96                                  & 91.82                                \\
\rowcolor[HTML]{EFEFEF} PViT + ED     & 2.72                                   & 99.15                                 & \textbf{25.29}                         & \textbf{93.87}                       & \textbf{33.63}                         & \textbf{92.29}                       & 42.38                                  & 89.07                                & \textbf{37.38}                         & 90.24                                & \textbf{24.93}                         & \textbf{94.67}                       & \textbf{52.01}                         & 86.12                                & \textbf{31.19}                         & \textbf{92.20}                       \\
\bottomrule
\end{tabular}%

}
    \caption{OOD detection results for PViT using the ViT-LP model with SWAG weights~\cite{singh2022revisiting} plus a linear classifier as the prior model. All benchmarks are evaluated on the same prior model with an ID accuracy of \textbf{80.66\%}, while PViT achieves an ID accuracy of \textbf{81.71\%}. OOD detection results are reported for \textsc{IMAGENET-1K} as the ID data. \textbf{Bold} numbers indicate superior results.}
    \label{tab:swag_imagenet}
\end{table*}

\vspace{2mm}
\noindent\textbf{Evaluation Metrics.} 
For assessing the performance of our proposed models in OOD detection, we employ two evaluation metrics: (1) \textbf{FPR95}, which measures the false positive rate of OOD samples when the true positive rate for ID samples is at 95\%; (2) \textbf{AUROC}, which computes the Area Under the Receiver Operating Characteristic Curve.

\subsection{Evaluation on OOD Detection} \label{sec:ood_detect}
We evaluate our \model's performance in OOD detection against competitive baselines including MSP~\cite{hendrycks2016baseline}, MaxLogit score~\cite{hendrycks2019scaling}, Mahalanobis score~\cite{lee2018simple}, Energy score~\cite{liu2020energy}, SSD~\cite{sehwag2021ssd}, ViM~\cite{wang2022vim}, KNN~\cite{sun2022out}, and NNGuide~\cite{park2023nearest}. To ensure fairness in comparison, we do not include any synthesis-based OOD methods, such as VOS~\cite{du2022towards} or Dream-OOD~\cite{du2023dream}, since our \model could be further also enhanced with the inclusion of synthesis data training. For all instances of \model shown in Tab.~\ref{tab:vit_imagenet} and Tab.~\ref{tab:swag_imagenet}, the scaling factor of the prior token \( \alpha \) is set to \( 0.1 \). 

As shown in Tab.~\ref{tab:vit_imagenet} and Tab.~\ref{tab:swag_imagenet}, \model is evaluated across a wide range of seven OOD datasets to demonstrate its superior performance. Our experiments show that \model exhibits remarkable performance on large-scale ID datasets \textsc{Imagenet-1k}, particularly when utilizing ViT-based prior models. Notably, even when compared to the Mahalanobis and ViM detectors, both of which are known for their strong performance in ViT-based architectures due to the Gaussian nature of the vision transformer embedding space~\cite{fort2021exploring,koner2021oodformer}, \model significantly outperforms these benchmarks in both the prior models of vanilla ViT variants.

\begin{table}
    \centering
    \scalebox{0.70}{

\begin{tabular}{lcccc}
\toprule
\multicolumn{1}{c}{Prior Model}                            & \multicolumn{2}{c}{ResNet18}        & \multicolumn{2}{c}{ViT}             \\ 
\cmidrule(lr){2-3} \cmidrule(lr){4-5} 
                                                           & FPR95$\downarrow$ & AUROC$\uparrow$ & FPR95$\downarrow$ & AUROC$\uparrow$ \\ 
\midrule
MSP      & 78.44             & 79.17           & 75.65             & 79.97           \\
MaxLogit & 76.88             & 80.90           & 54.65             & 87.45           \\
Energy   & 76.06             & 80.98           & 54.66             & 87.50           \\
SSD      & 87.77             & 63.47           & 50.03             & 89.39           \\
ViM      & 87.69             & 61.45           & \textbf{49.42}             & \textbf{89.49}           \\
KNN      & 99.72             & 42.87           & 99.67             & 40.07           \\
NNGuide  & \textbf{75.20}             & \textbf{81.39}           & 57.46             & 83.07           \\
\rowcolor[HTML]{EFEFEF} 
PViT + PCE                                                 & 79.77             & 79.25           & 65.04             & 85.42           \\
\rowcolor[HTML]{EFEFEF} 
PViT + KL                                                  & 85.20             & 76.96           & 76.02             & 80.73           \\
\rowcolor[HTML]{EFEFEF} 
PViT + ED                                                  & 81.90             & 77.94           & 66.22             & 83.66           \\ 
\bottomrule
\end{tabular}

}
    \caption{OOD detection results for \model with ResNet18 and ViT as prior models. The ID accuracy of ResNet18 and ViT on \textsc{CIFAR-100} is \textbf{77.27\%} and \textbf{85.80\%}. \model achieves ID accuracies of \textbf{78.84\%} and \textbf{81.78\%}, respectively. The average performance across six different OOD datasets is reported.}
    \label{tab:cifar100}
\end{table}   

To provide a comprehensive evaluation of \model, we also include results on the small-scale \textsc{CIFAR-100} dataset as the ID data in Tab.~\ref{tab:cifar100}. Although \model does not achieve superior performance compared to the baselines, which is likely due to the insufficient data samples in the small-scale \textsc{CIFAR-100} dataset and the need to scale image pixels from $32\times32$ to $224\times224$ to maintain consistency for the same configuration of \model, \model still ranks among the top performers. The results shown are the average over six OOD datasets.

\begin{figure*}[ht]
    \centering
    \begin{subfigure}[b]{0.33\textwidth}
        \centering
        \includegraphics[width=\textwidth]{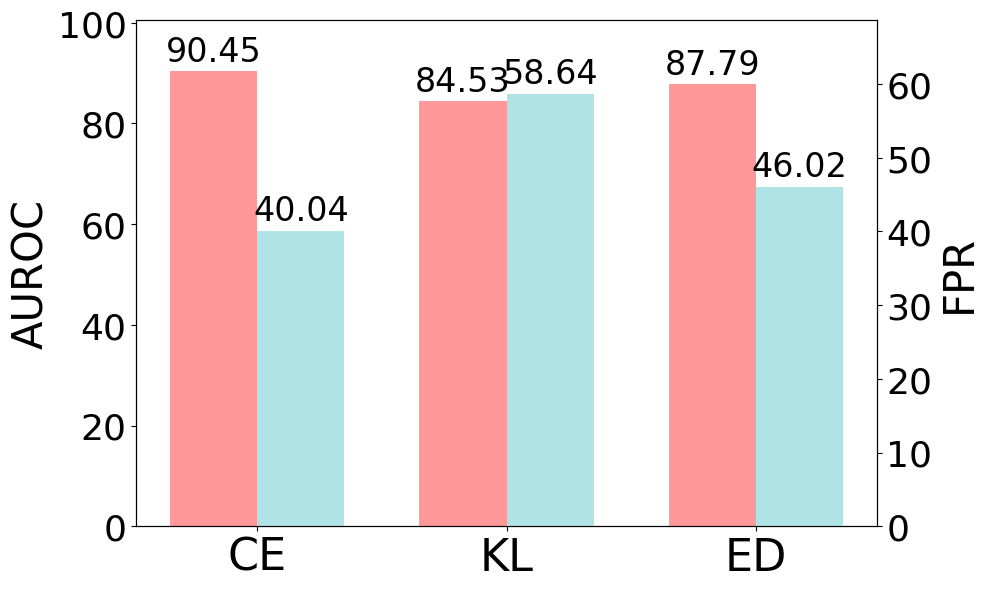}
        \caption{Results of ViT-B/16 as the prior model.}
    \end{subfigure}
    \hspace{10pt}
    \begin{subfigure}[b]{0.33\textwidth}
        \centering
        \includegraphics[width=\textwidth]{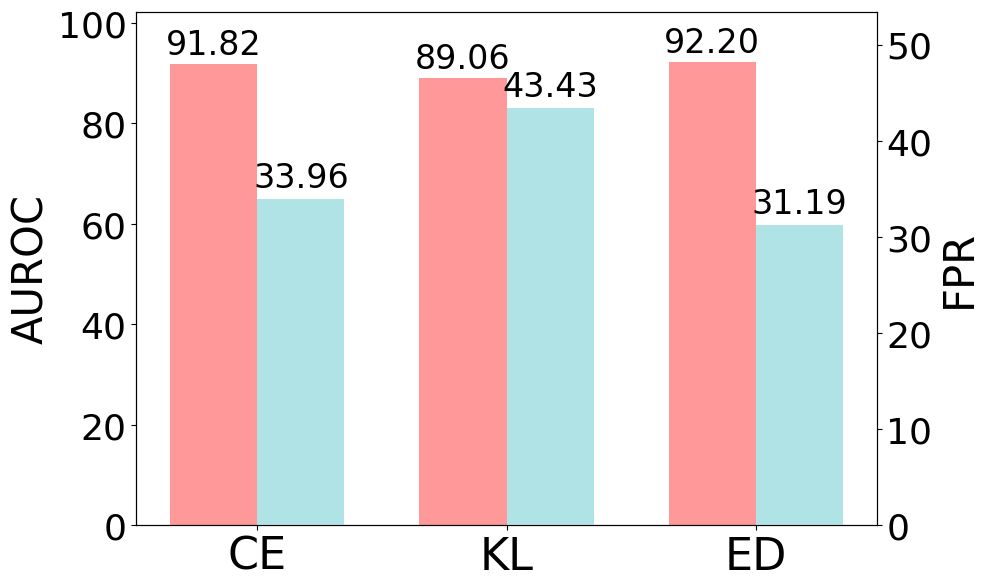}
        \caption{Results of using ViT-LP as the prior model.}
    \end{subfigure}    
    \caption{Ablation study on different scoring rules of \model. The ID data is \textsc{IMAGENET-1k}. The results are average results over seven OOD datasets.}
    \label{fig:aurocfpr}
\end{figure*}

\begin{figure*}[ht]
    \centering
    \includegraphics[width=0.65\textwidth]{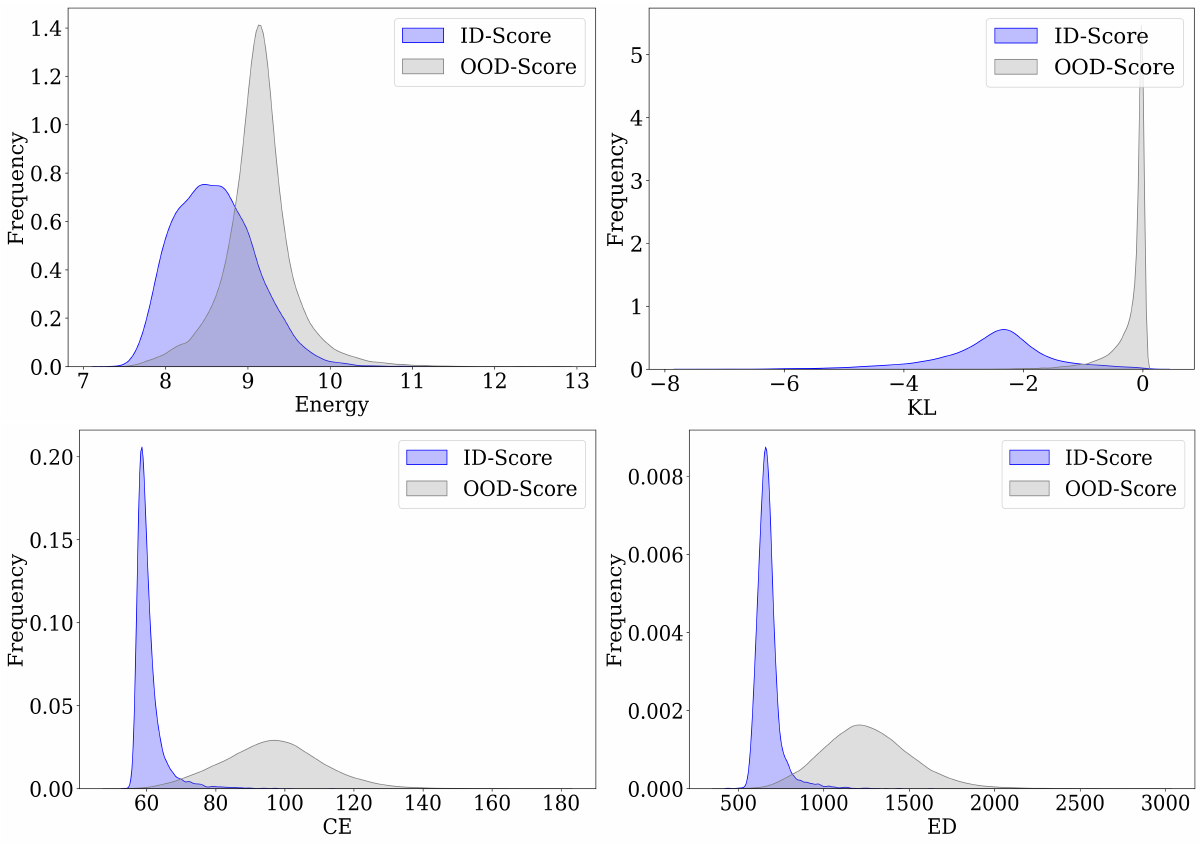}
    \caption{Score distributions with \textsc{IMAGENET-1K} as ID data and \textsc{iNaturalist} as OOD data. The scores are calculated by \model with ViT-LP as the prior model.}
    \label{fig:score}
\end{figure*}

\subsection{Ablation Studies} \label{sec:scale}
\noindent\textbf{Ablation Study on OOD Prior Guidance.} \label{sec:score} Although we have introduced using the CE as the guidance term in Eq.~\eqref{eq:guidance_term} for detecting OOD instances, we also considered other metrics to measure the difference between the priors and the predicted logits: 1) Euclidean Distance~\cite{danielsson1980euclidean} (ED). Euclidean Distance is a geometric measure calculating the "straight-line" distance between two points in Euclidean space. For vectors of predicted logits and prior probabilities, it is computed as \( \sqrt{\sum (p_i - q_i)^2} \), with \( p_i \) and \( q_i \) being the corresponding elements from the prior and predicted logits vectors, respectively. 2) Kullback–Leibler Divergence (KL-Divergence)~\cite{blundell2015weight}, used for measuring the distance between two probability distributions, is defined as \( \text{KL}(P \parallel Q) = \sum P(x) \log \frac{P(x)}{Q(x)} \), where \( P \) represents the true distribution of data (priors in our context) and \( Q \) denotes the distribution inferred by the model (predicted logits).


As visualized in Fig.~\ref{fig:score}, the score distributions reveal that our prior-guided energy score better distinguishes between ID and OOD data compared to the original energy score. Both CE and ED guidance terms produce similar results, albeit with different score values. The performance comparison in Fig.~\ref{fig:aurocfpr} further demonstrates that ED achieves performance comparable to that of the CE as the guidance.

\noindent\textbf{Ablation Study on Effect of Priors} 
To demonstrate the efficacy of our integrated prior token in guiding \model~to effectively differentiate between ID and OOD data, we conduct an ablation study comparing our approach with a vanilla ViT model. Specifically, we evaluate the OOD detection performance without prior knowledge integration by directly calculating the differences between a vanilla ViT model (provided by Google~\cite{dosovitskiy2020image}, details can be found in Appendix) and the prior models. The results are presented in Tab.~\ref{tab:noprior}.

\begin{table*}[ht]
    \centering
    \small
    \scalebox{0.55}{
    \begin{tabular}{llcccccccccccccccc}
\toprule
\multicolumn{2}{l}{OOD} & \multicolumn{2}{c}{\textsc{inaturalist}}   & \multicolumn{2}{c}{\textsc{sun}}           & \multicolumn{2}{c}{\textsc{places}}        & \multicolumn{2}{c}{\textsc{textures}}      & \multicolumn{2}{c}{\textsc{ninco}}         & \multicolumn{2}{c}{\textsc{openimage\_o}}  & \multicolumn{2}{c}{\textsc{SSB-hard}}      & \multicolumn{2}{c}{Mean}                                \\ 
\midrule
Prior model   & $G(\mathbf{x})$  & FPR95 & AUROC & FPR95 & AUROC & FPR95 & AUROC & FPR95 & AUROC & FPR95 & AUROC & FPR95 & AUROC & FPR95 & AUROC & FPR95 & AUROC \\ 
\midrule
ResNet50      & CE               & 100.00 & 12.32 & 100.00 & 19.81 & 100.00 & 20.13 & 100.00 & 16.93 & 100.00 & 18.72 & 100.00 & 14.44 & 100.00 & 29.51 & 100.00 & 18.84 \\
              & KL               & 94.45  & 79.78 & 94.08  & 76.27 & 94.09  & 76.04 & 84.63  & 80.64 & 90.23  & 79.10 & 90.82  & 81.42 & 94.06  & 69.37 & 91.77  & 77.52 \\
              & ED               & 74.32  & 81.70 & 70.20  & 81.95 & 69.20  & 81.63 & 84.59  & 73.63 & 83.48  & 72.86 & 82.58  & 74.65 & 85.25  & 63.70 & 78.52  & 75.73 \\ 
\midrule
vit-b-16      & CE               & 100.00 & 10.54 & 99.68  & 19.92 & 99.65  & 20.10 & 99.73  & 16.89 & 99.97  & 18.21 & 100.00 & 13.57 & 99.61  & 28.99 & 99.81  & 18.32 \\
              & KL               & 93.23  & 86.42 & 92.59  & 78.70 & 91.64  & 78.91 & 81.95  & 82.82 & 89.11  & 81.61 & 88.90  & 85.05 & 96.17  & 69.37 & 90.51  & 80.41 \\
              & ED               & 98.96  & 63.86 & 84.09  & 66.41 & 86.48  & 64.07 & 64.20  & 67.84 & 92.04  & 60.45 & 83.42  & 64.65 & 90.24  & 60.90 & 85.63  & 64.03 \\ 
\midrule
vit-lp        & CE               & 100.00 & 10.54 & 99.68  & 19.92 & 99.65  & 20.10 & 99.73  & 16.89 & 99.97  & 18.21 & 100.00 & 13.57 & 99.61  & 28.99 & 99.81  & 18.32 \\
              & KL               & 93.23  & 86.42 & 92.59  & 78.70 & 91.64  & 78.91 & 81.95  & 82.82 & 89.11  & 81.61 & 88.90  & 85.05 & 96.17  & 69.37 & 90.51  & 80.41 \\
              & ED               & 98.96  & 63.86 & 84.09  & 66.41 & 86.48  & 64.07 & 64.20  & 67.84 & 92.04  & 60.45 & 83.42  & 64.65 & 90.24  & 60.90 & 85.63  & 64.03 \\ 
\bottomrule
\end{tabular}
}
    \caption{Results of OOD detection using a vanilla ViT without prior tokens and a ViT with integrated prior tokens. ID dataset is \textsc{IMAGENET-1K}.}
    \label{tab:noprior}
\end{table*}

\begin{figure*}[ht]
    \centering
    \begin{subfigure}[b]{0.3\textwidth}
        \centering
        \includegraphics[width=\textwidth]{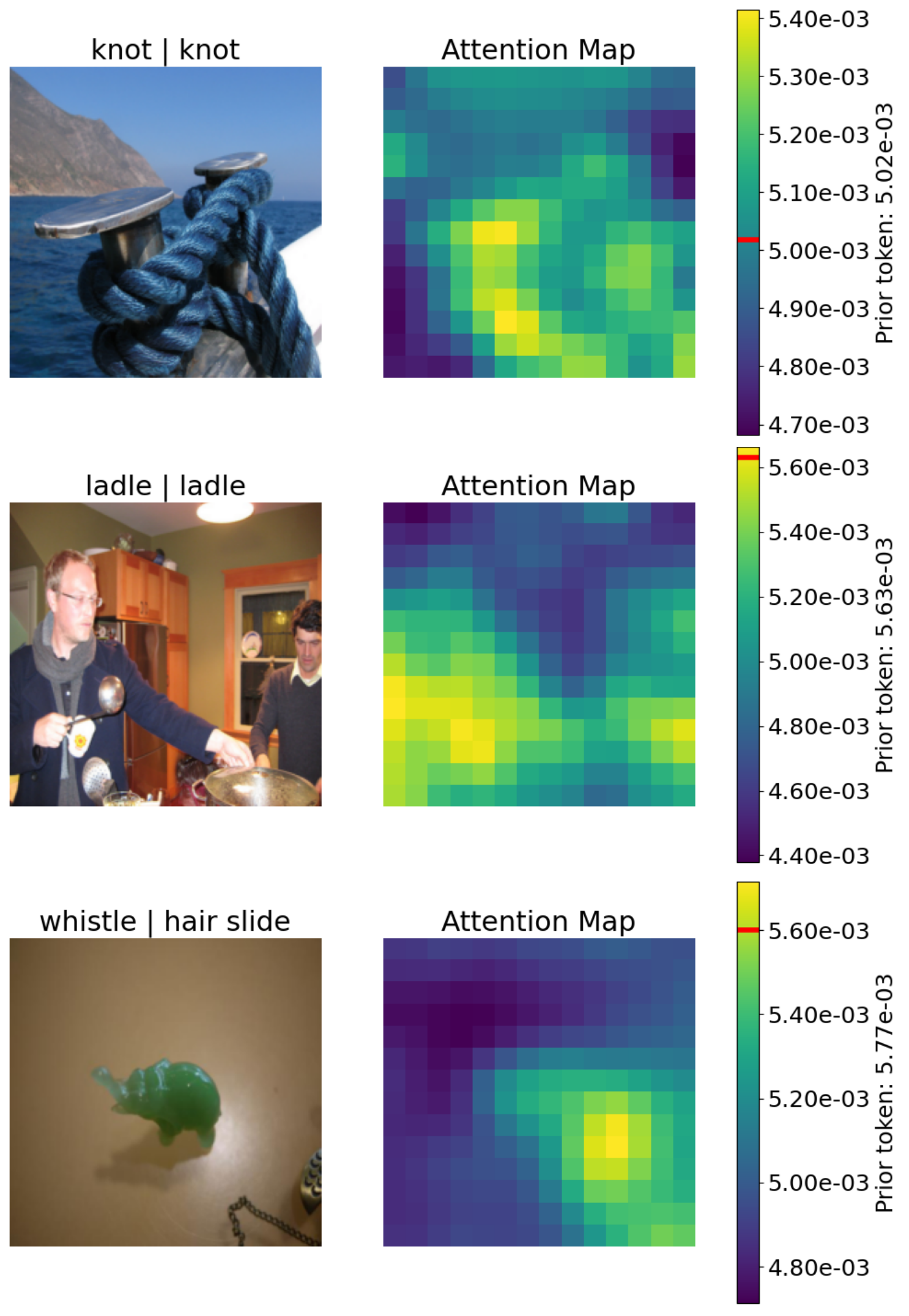}
        \caption{$\alpha = 10$.}
    \end{subfigure}
    \hfill
     \begin{subfigure}[b]{0.3\textwidth}
        \centering
        \includegraphics[width=\textwidth]{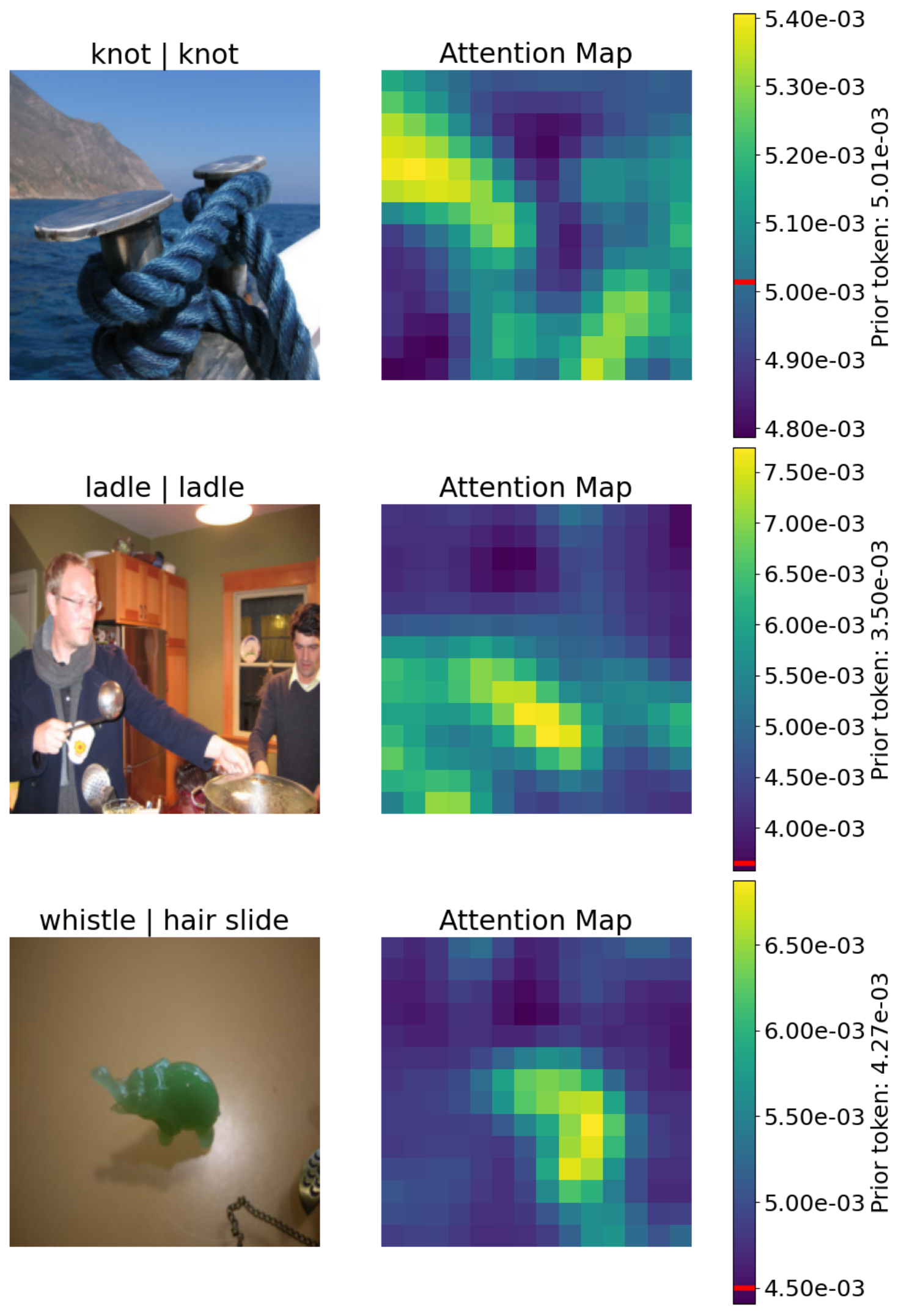}
        \caption{$\alpha = 1$.}
    \end{subfigure}
    \hfill
     \begin{subfigure}[b]{0.3\textwidth}
        \centering
        \includegraphics[width=\textwidth]{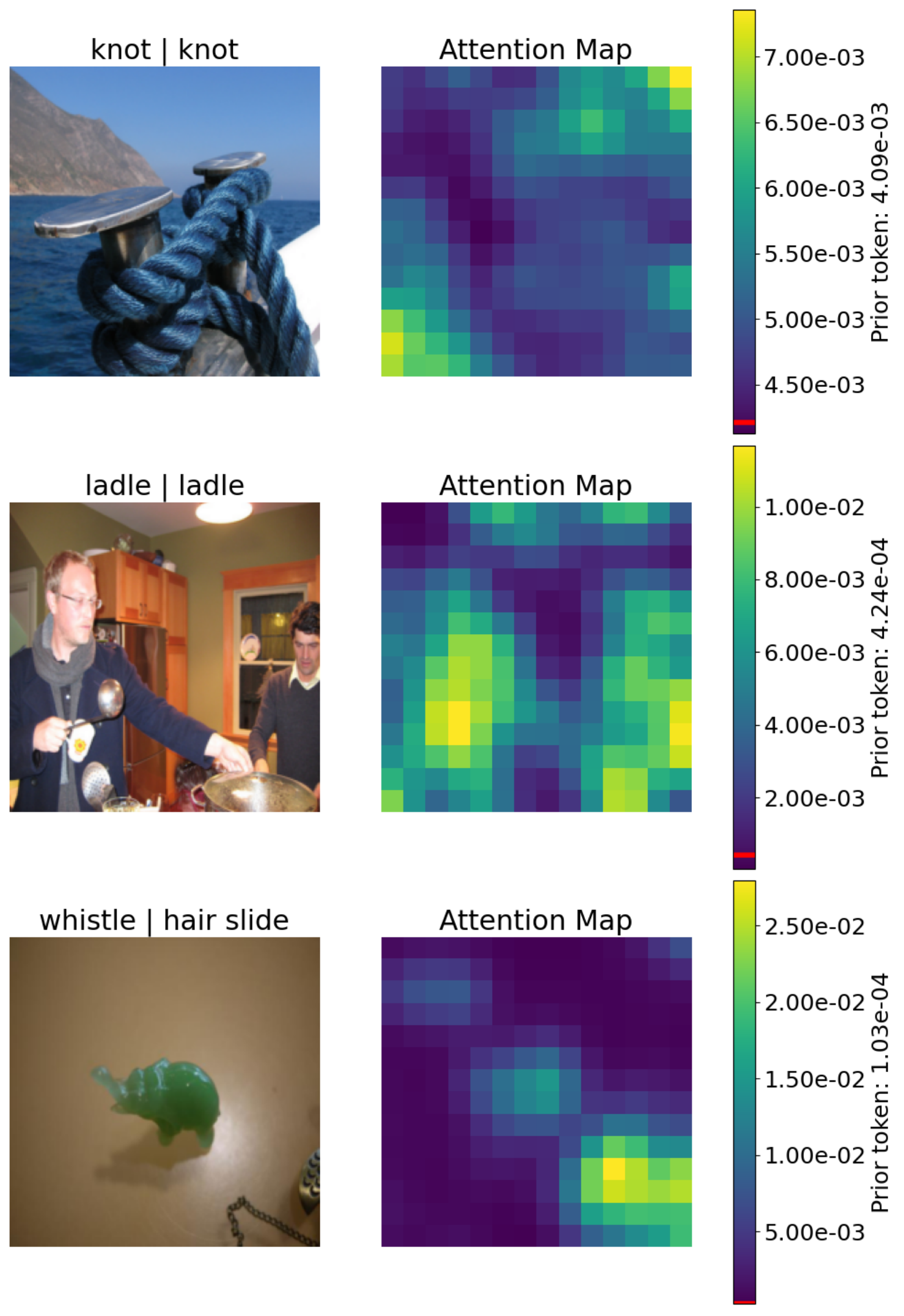}
        \caption{$\alpha = 0.1$.}
    \end{subfigure}
    \caption{Visualization of attention maps with varying scaling factors \( \alpha \) for prior token embedding, generated from the last layer and the first MSA head. The attention weight for the prior token is highlighted with a red line on the color bar. The first two rows of figures are taken from \textsc{IMAGENET-1K}, representing ID data. The third row, representing the OOD data, which is taken from \textsc{OpenImage\_O}, illustrates the differential responses of \model to both ID and OOD data. Labels above the original figure on the \textbf{left} indicate predictions made by the prior model, while the labels on the \textbf{right} correspond to the predictions made by \model.}
    \label{fig:atten_map}
\end{figure*}

As shown in Tab.~\ref{tab:noprior}, the performance of vanilla ViT is significantly inferior across all three types of prior guidance. Without the integration of prior tokens, OOD detection using CE as prior guidance achieves nearly 100\% FPR95 on the mean of seven OOD datasets. Similarly, the results obtained with KL and ED as prior guidance are also suboptimal, with FPR95 and AUROC metrics performing poorly compared to the results presented in Tab.~\ref{tab:vit_imagenet} and Tab.~\ref{tab:swag_imagenet}. These findings underscore that the ViT architecture, in its vanilla form, lacks the inherent capacity to effectively differentiate between OOD and ID data. This demonstrates that the integration of prior tokens is crucial for PViT to achieve robust OOD detection.

\noindent\textbf{Ablation Study on Scaling the Priors.} Fig.~\ref{fig:atten_map} illustrates the impact of scaling prior weights \( \alpha \) on \model's output. Our analysis aims to mitigate the excessive influence of priors. It is observed that a higher \( \alpha \) leads to more attention on the prior token, potentially reducing the focus on image patches. Conversely, a lower \( \alpha \) may enhance attention on image patches. Optimal performance is achieved when \model balances its focus between the prior token and image tokens. This balance results in a clearer distinction between priors and predictions, underpinning the effectiveness of \model in OOD detection. 


To further investigate the trade-offs in attention weights among patch tokens, we visualize the attention map in Fig.~\ref{fig:atten_map}, using image examples with corresponding prior and predicted labels. The color bar in these maps ranges from highest to lowest attention, highlighting the impact of the scaling factor \( \alpha \) on the model's attention mechanism. Notably, the attention weight for the prior token, indicative of \model's focus on the prior token, is demarcated with a red line on the color bar. As a result of OOD detection, we can see that in the third row of figures, \model often produces predictions that notably differ from the prior model's.


\section{Discussions and Future Work} \label{sec:diss}
\vspace{2mm}
\subsection{Discussion from Bayesian Perspective} Bayesian Neural Networks (BNNs) have been explored for OOD detection in various studies~\cite{d2021out, pmlr-v80-wang18i, gal2016dropout, wang2021bayesian, 10224150, DBLP:journals/corr/abs-2107-12248}. BNNs, which integrate Bayesian methods into neural networks, utilize probability distributions over model parameters to represent uncertainties in predictions~\cite{neal1995bayesian}. In the context of OOD detection, BNNs can be employed by comparing uncertainties between the model's predictions on given inputs and known ID data. However, the suitability of BNNs for OOD detection has been a subject of debate in recent works~\cite{DBLP:journals/corr/abs-2107-12248}.

From a Bayesian perspective, our approach can be interpreted as utilizing prior knowledge from an ID dataset to establish the Predictive Posterior Distribution (PPD) within a Bayesian framework. This aligns with the concept of Transformers facilitating Bayesian inference~\cite{müller2023transformers}. In our model, the priors can be treated akin to mean values of sampled priors from a Bayesian model~\cite{10224150}, positioning \model to approximate the posterior distribution \( P(y|x, \mathcal{D}, \pi) \). This approximation follows the equation \( p(y|x,\mathcal{D}) = \int_t p(y|x,t)p(t|\mathcal{D}) \), where \( P(x | y, \mathcal{D}, \pi) \) denotes the likelihood of observing the data given the label and priors, \( P(y | \mathcal{D}, \pi) \) represents the prior probability informed by the dataset and the prior model, and \( P(x | \mathcal{D}, \pi) \) signifies the evidence, usually computed by marginalizing over the label space. However, the method proposed by~\cite{müller2023transformers} is specifically designed for single-sequence data, aimed at providing ultra fast Bayesian inference in a single forward pass~\cite{hollmann2023tabpfn}. This approach, while not directly applicable to image data due to ViTs' processing limitations, sheds light on the efficacy of our \model in OOD detection. By capturing uncertainties through Bayesian inference, it provides a compelling explanation for \model's robust performance in identifying OOD samples.

\subsection{Discussion of Inductive Bias} Gernarally, ViTs and CNNs are believed that they exhibit fundamentally different inductive biases~\cite{dosovitskiy2020image, li2021localvit, jelassi2022vision}. ViTs inherently focus on global image patterns by treating image patches as analogous to tokens. This global perspective contrasts sharply with the local feature emphasis of CNNs, which inherently encode a bias towards local spatial hierarchies and proximities. While this enables ViTs to excel in tasks requiring holistic image comprehension, their lack of built-in locality bias may limit effectiveness in tasks where detailed local feature analysis is crucial~\cite{xu2021vitae, jelassi2022vision}. 

ViTs can adopt inductive biases through data augmentation or hybrid architectures, improving their local feature processing, traditionally a strength of CNNs~\cite{xu2021vitae, jelassi2022vision}. Our study introduces embedding additional prior information into ViTs, enhancing their robustness and serving as a method to incorporate inductive bias. This strategy marks a new path for embedding manually designed inductive biases into ViTs, potentially boosting their robustness and explainability. By introducing additional prior information, we hope our \model can compensate for the less pronounced traditional inductive biases in ViTs, where such augmentation can significantly refine their performance, particularly in challenging scenarios where inductive biases play a crucial role.


\subsection{Discussion and Future Work} The accuracy of the proposed approach is related with the accuracy and structure of the prior model, especially on the ID prior accuracy. PViT shows better OOD detection when using ViT architectures as priors compared to CNNs, likely due to structural similarities. Additionally, PViT requires training on ID data, which, despite benefiting from prior knowledge and achieving rapid convergence, introduces extra complexity. During inference, the need to perform inference on both the prior model and PViT also increases computational cost.

Furthermore, our exploration reveals that \model introduces a beneficial inductive bias to large vision models, which are increasingly prevalent and continuously evolving. Similarly to the advancements witnessed in Large Language Models (LLMs)~\cite{chang2023survey}, the field of computer vision has also seen a significant proliferation of large vision models~\cite{kirillov2023segany,bai2023sequential}. These large models are facing the challenge of "planning"~\cite{NEURIPS2023_7a92bcde}—a critical aspect of directing these models' capabilities towards specific, controlled outcomes. Looking toward the future, expanding \model to incorporate varying levels of priors could direct large vision models towards specific objectives, akin to customizing OOD detection for diverse scenarios. This represents a promising research direction with potential implications across different vision transformer architectures~\cite{liu2021swin}.



\section{Conclusion} \label{sec:conclusion}
In this work, we present Prior-augmented Vision Transformer, a novel and generic framework for OOD detection. \model uniquely integrates prior knowledge as a prior token to be trained to approximate the true label, allowing for effective differentiation of OOD data by examining the relative distances between model predictions and the prior logits. Our empirical results demonstrate that \model achieves outstanding performance in OOD detection benchmarks. Moreover, the innovative integration of prior knowledge by \model not only enhances OOD detection capabilities but also suggests a versatile approach for the strategic planning and control of large vision models tailored to specific practical applications.\\ 


\noindent \textbf{Acknowledgements.} We are grateful to the UK EPSRC through
Project NSF-EPSRC: ShiRAS. Towards Safe and Reliable Autonomy in Sensor
Driven Systems, under Grant EP/T013265/1, and by the USA National Science Foundation under Grant NSF ECCS 1903466. This work was also supported by the UKRI Trustworthy Autonomous Systems Node in Resilience (REASON) 
EP/V026747/1 project.

\bibliographystyle{elsarticle-num}
\bibliography{main}
\clearpage
\setcounter{page}{1}
\appendix

\renewcommand\thefigure{\Alph{section}\arabic{figure}}
\renewcommand\thetable{\Alph{section}\arabic{table}}
\renewcommand{\thethm}{\arabic{thm}}
\setcounter{figure}{0}
\setcounter{table}{0}
\setcounter{thm}{0}

\section{Implementation Details for Reproducibility}
\subsection{ID datasets} \label{appen:data}
\vspace{2mm}  \noindent\textbf{CIFAR10} consists of 60,000 32x32 color images in 10 classes, with 6,000 images per class~\cite{cifar}. The dataset is divided into 50,000 training images and 10,000 test images. The classes are mutually exclusive and include objects such as cats, dogs, trucks, and ships.

\vspace{2mm}  \noindent\textbf{CIFAR100} is similar to CIFAR10 but contains 100 classes each consisting of 600 images~\cite{cifar}. Each class is divided into 500 training images and 100 testing images. The 100 classes in CIFAR100 are grouped into 20 super-classes, each comprising 5 sub-classes.

\vspace{2mm}  \noindent\textbf{ImageNet-1K}, a subset of the larger ImageNet dataset, contains over 1.2 million images spanning 1,000 classes~\cite{deng2009imagenet}. \textsc{ImageNet-1K} is a standard benchmark in computer vision research. Its classes encompass a wide range of objects, including various species of animals, plants, and everyday objects.

\subsection{OOD datasets} 
\vspace{2mm}  \noindent\textbf{Tiny ImageNet} is derived from the larger ImageNet dataset, scaled down to include 200 classes. Each class contains 500 training images, 50 validation images, and 50 test images~\cite{zhang2023openood}. This dataset serves as a more compact yet challenging benchmark for image classification.

\vspace{2mm}  \noindent\textbf{Textures} comprises diverse images of textures categorized into several classes, providing a unique challenge for texture recognition and classification models~\cite{DBLP:conf/cvpr/CimpoiMKMV14}. It is widely used for evaluating model robustness against textural variations.

\vspace{2mm}  \noindent\textbf{SVHN (Street View House Numbers)} contains digit images obtained from house numbers in Google Street View images. It includes over 600,000 digit images, making it a comprehensive dataset for digit classification tasks~\cite{netzer2011reading}.

\vspace{2mm}  \noindent\textbf{Places} is a large-scale dataset of scene-centric images. With 365 scene categories and over 1.8 million images, it is extensively used for scene recognition and contextual understanding in images~\cite{DBLP:journals/pami/ZhouLKO018}.

\vspace{2mm}  \noindent\textbf{Sun} is a dataset of natural scene images under varying illumination and weather conditions, often used for assessing model performance in diverse environmental settings~\cite{DBLP:journals/corr/XuEZFKX15}.

\vspace{2mm}  \noindent\textbf{LSUN (Large-scale Scene UNderstanding)} consists of around one million labeled images for each of 10 scene categories and 20 object categories, providing a diverse set for scene understanding and object detection tasks~\cite{DBLP:journals/corr/YuZSSX15}.

\vspace{2mm}  \noindent\textbf{SSB-hard} and \textbf{NINCO} are datasets specifically designed for evaluating OOD detection in neural networks. SSB-hard focuses on subtly different classes, while NINCO provides near-in-context OOD examples, presenting unique challenges for OOD detection~\cite{vaze2022openset, bitterwolf2023ninco}.

\vspace{2mm}  \noindent\textbf{OpenImage-O} is a subset of the OpenImages dataset, tailored for OOD detection. It includes a wide range of object categories not present in standard datasets like ImageNet, making it ideal for testing the generalizability of models~\cite{haoqi2022vim}.

\vspace{2mm}  \noindent\textbf{iNaturalist} contains images of natural world~\cite{van2018inaturalist}. It has 13 super-categories and 5,089 sub-categories covering plants, insects, birds, mammals, and so on. We use the subset that contains 110 plant classes which do not overlap with \textsc{Imagenet-1k}.

\subsection{Prior Models}
\label{appen:prior}

\vspace{2mm} \noindent\textbf{Pretrained ResNet-18 on CIFAR10:} Achieving a test accuracy of 0.9498, this model was trained with a batch size of 128 over 300 epochs, and validation every 5 epochs. The SGD optimizer was used with a learning rate of 0.1, momentum of 0.9, and weight decay of 0.0005, paired with a ReduceLROnPlateau scheduler. This pretrained ResNet-18 model is available on Hugging Face.\footnote{\url{https://huggingface.co/edadaltocg/resnet18_cifar10}}

\vspace{2mm} \noindent\textbf{Pretrained ViT on CIFAR10:} The model, with a test accuracy of 0.9788 and a loss of 0.2564, was trained using an Adam optimizer with a learning rate of 5e-05, and a linear learning rate scheduler. This model can be found on Hugging Face.\footnote{\url{https://huggingface.co/aaraki/vit-base-patch16-224-in21k-finetuned-cifar10}}

\vspace{2mm} \noindent\textbf{Pretrained ResNet-18 on CIFAR100:} This model recorded a test accuracy of 0.7926. Using SGD as the optimizer with a learning rate of 0.1, momentum of 0.9, and weight decay of 0.0005, along with a CosineAnnealingLR scheduler, the model is available on Hugging Face.\footnote{\url{https://huggingface.co/edadaltocg/resnet18_cifar100}}

\vspace{2mm} \noindent\textbf{Pretrained ViT on CIFAR100:} The fine-tuned version of google/vit-base-patch16-224-in21k on CIFAR100 achieved an accuracy of 0.8985 and a loss of 0.4420. It used a learning rate of 0.0002, with train and eval batch sizes of 16 and 8. The pretrained ViT is available at Hugging Face.\footnote{\url{https://huggingface.co/Ahmed9275/Vit-Cifar100}}

\vspace{2mm} \noindent\textbf{Pretrained ResNet-50 on IMAGENET-1K:} Provided by Microsoft, this model achieved an accuracy of approximately 67.35\%. The model can be accessed at Hugging Face. \footnote{\url{https://huggingface.co/microsoft/resnet-18}}

\vspace{2mm} \noindent\textbf{Pretrained Google ViT on IMAGENET-1K:} The pretrained orginal ViT, fine-tuned on ImageNet-1K, was initially trained on ImageNet-21k. This model is available at Hugging Face. \footnote{\url{https://huggingface.co/google/vit-base-patch16-224}}. 

\vspace{2mm} \noindent\textbf{Pretrained ViT Variants on ImageNet-1K:} We explore various ViT configurations as the prior models, including models trained with different approaches. These configurations include weights trained using the DeIT training recipe, as well as models with the original frozen SWAG trunk weights combined with a linear classifier. The models are avaible in PyTorch. \footnote{\url{https://pytorch.org/vision/main/models/generated/torchvision.models.vit_b_16.html##torchvision.models.vit_b_16}}.

\end{document}